\documentclass[sigconf, nonacm]{acmart}

\newcommand\vldbdoi{XX.XX/XXX.XX}
\newcommand\vldbpages{XXX-XXX}
\newcommand\vldbvolume{14}
\newcommand\vldbissue{1}
\newcommand\vldbyear{2020}
\newcommand\vldbauthors{\authors}
\newcommand\vldbtitle{\shorttitle} 
\newcommand\vldbavailabilityurl{URL_TO_YOUR_ARTIFACTS}
\newcommand\vldbpagestyle{plain} 

\usepackage{multirow}
\usepackage{mathtools}
\usepackage{tcolorbox}
\tcbuselibrary{skins,breakable}
\usepackage{graphicx}
\usepackage{booktabs}
\usepackage{amsfonts} 
\usepackage{pgf-pie}
\usepackage{algorithm}
\usepackage{algpseudocode}

\begin{document}
\title{Ontology-Guided Evidence Path Inference for Multi-hop Knowledge Graph Question Answering}

\author{Yongxue Shan}
\affiliation{
  \institution{National University of Defense Technology}
  \city{Changsha}
  \country{China}
}
\email{shanyongxue001@nudt.edu.cn}

\author{Meihan Wu}
\affiliation{%
  \institution{Pengcheng Laboratory}
  \city{Shenzhen}
  \country{China}
}
\email{wumh@pcl.ac.cn}

\author{Cundi Fang}
\affiliation{%
  \institution{National University of Defense Technology}
  \city{Changsha}
  \country{China}
}
\email{fangcundi@nudt.edu.cn}

\author{Jie Peng}
\affiliation{%
  \institution{National University of Defense Technology}
  \city{Changsha}
  \country{China}
}
\email{pengjie@nudt.edu.cn}

\author{Xiaodong Wang}
\affiliation{%
  \institution{National University of Defense Technology}
  \city{Changsha}
  \country{China}
}
\email{xdwang@nudt.edu.cn}


\begin{abstract}

Knowledge graph question answering (KGQA) aims to answer natural-language questions by reasoning over structured facts. Existing multi-hop KGQA methods mainly rely on topic-centered expansion, which faces two key challenges: the search space rapidly grows with noisy mixed-type paths, and retrieved paths may fail to satisfy the semantic constraints of complex questions. To address these challenges, we propose OPI, an ontology-guided evidence path inference framework for multi-hop KGQA. OPI introduces a relation-centric ontology graph to capture the head-tail type constraints of relations, providing a compact interface for answer-side constraints. Based on this ontology graph, OPI first introduces a bidirectional retrieval mechanism by mapping the predicted answer type to compatible final-hop relations and combining topic-side prefix expansion with answer-side final-hop matching, thereby suppressing noisy mixed-type expansion. OPI further adopts an iterative refinement strategy to reassess retrieved paths and candidate answers under the question context, filtering type-compatible but question-irrelevant evidence for more reliable answer prediction. Experiments on WebQSP, CWQ, and MetaQA show that OPI substantially reduces the search space, improves Hit@1/F1 by 4.6/5.0 points on WebQSP and 8.9/3.3 points on CWQ over the strongest prior results, and achieves near-saturated Hit@1 on MetaQA with the retrieval module alone.
\end{abstract}

\maketitle

\pagestyle{\vldbpagestyle}
\begingroup\small\noindent\raggedright\textbf{PVLDB Reference Format:}\\
\vldbauthors. \vldbtitle. PVLDB, \vldbvolume(\vldbissue): \vldbpages, \vldbyear.\\
\href{https://doi.org/\vldbdoi}{doi:\vldbdoi}
\endgroup
\begingroup
\renewcommand\thefootnote{}\footnote{\noindent
This work is licensed under the Creative Commons BY-NC-ND 4.0 International License. Visit \url{https://creativecommons.org/licenses/by-nc-nd/4.0/} to view a copy of this license. For any use beyond those covered by this license, obtain permission by emailing \href{mailto:info@vldb.org}{info@vldb.org}. Copyright is held by the owner/author(s). Publication rights licensed to the VLDB Endowment. \\
\raggedright Proceedings of the VLDB Endowment, Vol. \vldbvolume, No. \vldbissue\ %
ISSN 2150-8097. \\
\href{https://doi.org/\vldbdoi}{doi:\vldbdoi} \\
}\addtocounter{footnote}{-1}\endgroup

\ifdefempty{\vldbavailabilityurl}{}{
\vspace{.3cm}
\begingroup\small\noindent\raggedright\textbf{PVLDB Artifact Availability:}\\
The source code, data, and/or other artifacts have been made available at \url{https://github.com/shanyongxue/OPI_KGQA}.
\endgroup
}

\section{Introduction}
Knowledge graph question answering (KGQA) aims to answer natural-language questions by reasoning over entities and relations in a knowledge graph, and has been applied in domains such as healthcare \citep{frisoni2024generate}, agriculture \citep{yang2024agriqa}, and multimedia \citep{lee2024mrmkg}. Driven by the rapid advances of large language models (LLMs), recent studies increasingly integrate LLMs with KGs to support question understanding, reasoning planning, and graph-grounded answer generation \citep{pan2024unifying}. These methods explore diverse strategies, including iterative graph exploration, relation-path planning, and graph-constrained decoding, and have advanced multi-hop reasoning over KGs \citep{liu2025ort,wen2024mindmap,sun2024tog,luo2024rog,luo2025gcr,tan2025pog,mavromatis2025gnnrag}. Nevertheless, complex multi-hop KGQA remains challenging over large-scale KGs, where increasing scale, heterogeneity, and structural complexity introduce large search spaces and ambiguous reasoning evidence \citep{dong2023generations,ge2021largeea,rabbani2023extraction}.


A common paradigm for multi-hop KGQA is to retrieve evidence paths or subgraphs rooted at the topic entity and use them to derive answers. To improve retrieval efficiency, many methods control the expansion space through local subgraph construction, iterative graph exploration, and path pruning \citep{wen2024mindmap,sun2024tog,tan2025pog}. To improve reasoning quality, recent methods further introduce semantic signals such as relation-path planning, graph-structured prompting, and LLM-based evidence reasoning \citep{liu2025ort,luo2024rog,luo2025gcr,mavromatis2025gnnrag}. Despite these advances, most retrieval procedures are still primarily driven by topic-centered expansion: they start from the topic entity and progressively explore neighboring entities and relations. Consequently, retrieval often produces numerous structurally reachable paths that are inconsistent with the expected answer type, while also introducing spurious candidate paths that fail to satisfy the complex semantic constraints of the question.

\begin{figure}[t]
  \centering
  \includegraphics[width=\linewidth]{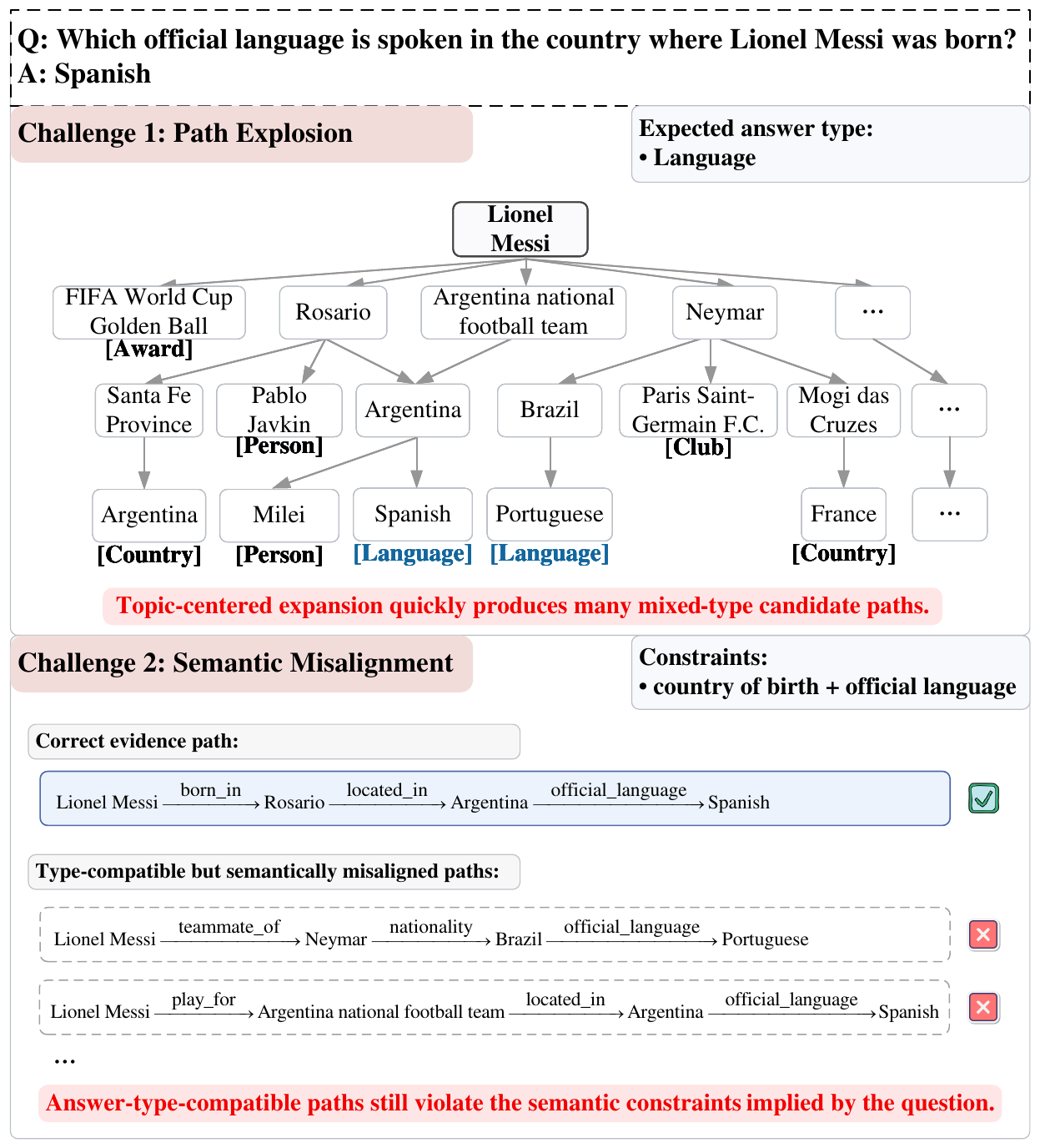}
  \caption{Illustration of two main challenges in multi-hop KGQA. The upper part shows that topic-centered expansion produces many mixed-type candidate paths, while the lower part shows that even answer-type-compatible paths may still violate the semantic constraints implied by the question.}
  \label{fig:challenge}
\end{figure}

This topic-centered retrieval paradigm faces two challenges in multi-hop KGQA, as illustrated in \autoref{fig:challenge}. The first is \textit{path explosion}: unconstrained forward expansion from the topic entity produces a massive expansion of candidate paths ending in heterogeneous types, the vast majority of which are completely irrelevant to the expected answer type. As shown in the upper part of \autoref{fig:challenge}, paths starting from the same topic entity may branch into countries, persons, awards, clubs, languages, and other mixed-type endpoints. The second is \textit{semantic misalignment}: even when candidate paths reach answer-type-compatible entities, they may still violate the implicit constraints of the question and thus deviate from the intended reasoning semantics. As shown in the lower part of \autoref{fig:challenge}, several paths can lead to plausible language entities, but only the path that captures both the birth-country constraint and the official-language constraint provides the correct reasoning evidence. Therefore, an effective multi-hop KGQA framework requires the ability to both mitigate the explosion of candidate paths and identify reasoning chains that satisfy the complex semantic constraints of the question.

To address these challenges, we propose OPI, an \textbf{O}ntology-guided evidence \textbf{P}ath \textbf{I}nference framework for multi-hop KGQA. Rather than relying on unconstrained topic-centered expansion that produces noisy, mixed-type paths, OPI introduces a relation-centric ontology graph as the structural basis for evidence path inference. This graph abstracts the knowledge graph into a type-level representation that captures how relations connect head and tail entity types, providing a compact interface for answer-side constraints. 
Driven by this compact interface, OPI replaces unconstrained topic-centered retrieval with a two-stage inference paradigm that addresses the two challenges. It first leverages a bidirectional retrieval mechanism to impose answer-side constraints and prune the explosive search space early. It then applies an iterative refinement strategy to filter spurious evidence and improve semantic alignment with the question.




Specifically, OPI leverages \textit{a bidirectional retrieval mechanism} to obtain a set of answer-type-compatible evidence paths. Given a question, OPI predicts the implied answer type and maps it to a small set of answer-type-compatible final-hop relations. It then combines topic-side prefix expansion with answer-side final-hop matching, so that retrieval is no longer driven by unconstrained topic-centered expansion alone. This bidirectional retrieval mechanism effectively suppresses mixed-type path growth and mitigates structural path explosion. After retrieval, OPI applies \textit{an iterative answer refinement strategy} to reassess the retrieved paths and candidate answers under the question context. The generator produces an answer hypothesis from the current path and answer contexts, while the refiner evaluates the hypothesis together with the retrieved evidence and returns structured feedback. This feedback updates the focused path context and candidate answer context, enabling OPI to filter type-compatible but question-irrelevant evidence and generate more reliable final answers.

Our main contributions are summarized as follows:
\begin{itemize}
    \item We introduce a relation-centric ontology graph, which captures how relations connect head and tail entity types and provides the structural basis for answer-side constrained path retrieval.
    \item We propose an ontology-guided bidirectional retrieval mechanism, which combines topic-side prefix expansion with answer-side final-hop matching to reduce noisy mixed-type expansion and alleviate path explosion.
    \item We design an iterative answer refinement strategy, which jointly reassesses retrieved paths and candidate answers under the question context to filter type-compatible but question-irrelevant evidence.
    \item We conduct extensive experiments on WebQSP, CWQ, and MetaQA, demonstrating the effectiveness of OPI and its key components across different KGQA benchmarks.
    
\end{itemize}

\section{Preliminaries}
In this section, we formalize the KGQA task considered in this paper by defining the knowledge graph, its type-level abstraction, and the answer prediction objective.

\subsection{Knowledge Graph and Ontology Graph}

\textbf{Knowledge graph.} 
A knowledge graph is denoted by $\mathcal{G}=(\mathcal{E},\mathcal{R},\mathcal{T})$, where:
\begin{itemize}
    \item $\mathcal{E}$ is the set of entities;
    \item $\mathcal{R}$ is the set of relations;
    \item $\mathcal{T}\subseteq \mathcal{E}\times\mathcal{R}\times\mathcal{E}$ is the set of factual triples.
\end{itemize}
Each factual triple $(e_h,r,e_t)\in\mathcal{T}$ indicates that the head entity $e_h$ is connected to the tail entity $e_t$ by relation $r$.

\begin{figure}[t]
  \centering
  \includegraphics[width=\linewidth]{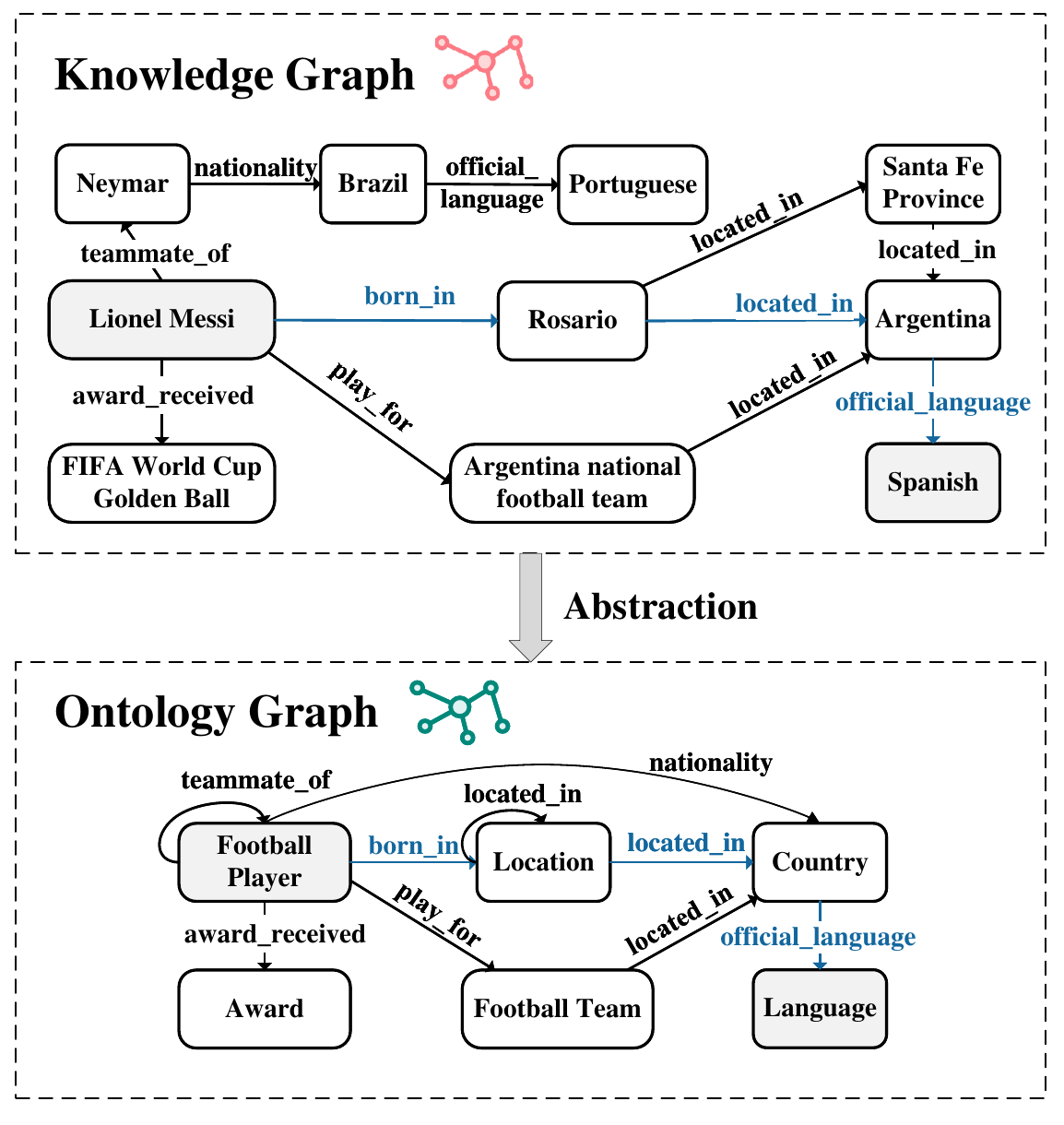}
  \caption{An example of a knowledge graph and its ontology graph. The ontology graph abstracts entity-level triples into type-level relation signatures.}
  \label{fig:kg_ontology}
\end{figure}

\textbf{Ontology graph.}
The ontology graph provides a type-level abstraction of the knowledge graph through relation signatures.
It is denoted by $\mathcal{O}=(\mathcal{C},\mathcal{R},\mathcal{S})$, where:
\begin{itemize}
    \item $\mathcal{C}$ is the set of semantic types associated with entities;
    \item $\mathcal{R}$ is the relation set shared with the knowledge graph;
    \item $\mathcal{S}\subseteq \mathcal{C}\times\mathcal{R}\times\mathcal{C}$ is the set of type-level relation signatures.
\end{itemize}
Each relation signature $(c_h,r,c_t)\in\mathcal{S}$ specifies that relation $r$ can connect head entities of type $c_h$ to tail entities of type $c_t$ at the ontology level. 
While the knowledge graph stores concrete entity-level facts, the ontology graph captures type-level compatibility through these relation signatures. 
\autoref{fig:kg_ontology} illustrates how entity-level triples in the knowledge graph are abstracted into type-level relation signatures in the ontology graph.

\subsection{Knowledge Graph Question Answering}
Given a natural language question $q$, we assume that a topic entity $e_q\in\mathcal{E}$ mentioned in the question is identified following standard KGQA settings~\citep{jiang2023unikgqa,luo2024rog}.
An evidence path rooted at $e_q$ is a sequence of connected factual triples:
\begin{equation}
\label{eq:reasoning_path}
p=(e_0 \xrightarrow{r_1} e_1 \xrightarrow{r_2} \cdots \xrightarrow{r_l} e_l),
\end{equation}
where $e_0=e_q$ and $(e_{i-1},r_i,e_i)\in\mathcal{T}$ for each $1\leq i\leq l$.
The terminal entity $e_l$ of an evidence path is regarded as a candidate answer grounded in the knowledge graph.

For each question, the ground-truth answer set is denoted by $\mathcal{Y}^*\subseteq\mathcal{E}$.
The task is to learn a function
\begin{equation}
\label{eq:kgqa_objective}
f_\theta(q,e_q,\mathcal{G},\mathcal{O}) \rightarrow \hat{y},
\end{equation}
where $\hat{y}\in\mathcal{E}$ denotes the final predicted answer. We assume that each correct answer $y\in\mathcal{Y}^*$ can be reached from the topic entity through at least one bounded-length evidence path in $\mathcal{G}$.
The objective is therefore to produce an answer prediction $\hat{y}$ that is graph-reachable from the topic entity and semantically consistent with the constraints expressed by the question.

\section{Method}

\subsection{Overall Architecture}
\begin{figure*}
  \centering
  \includegraphics[width=0.9\linewidth]{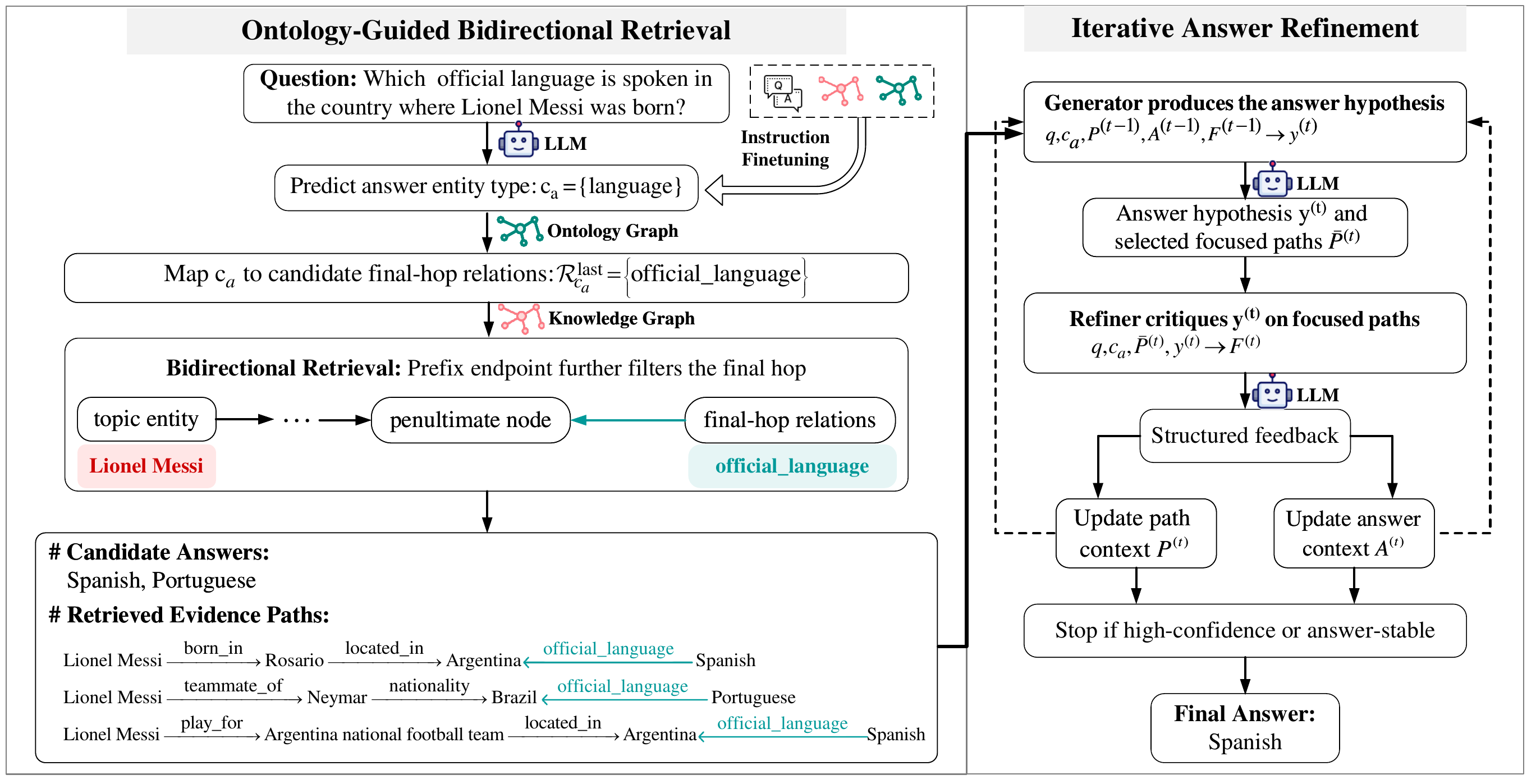}
    \caption{Overall inference pipeline of OPI. The left part shows ontology-guided bidirectional retrieval, which retrieves tail-type-compatible evidence paths. The right part shows iterative answer refinement, where a generator--refiner loop refines these paths and candidate answers to produce the final answer.}
  \label{fig:overview_part2}
\end{figure*}

OPI consists of three tightly connected modules: ontology graph construction, ontology-guided bidirectional retrieval, and iterative answer refinement. It first abstracts the original knowledge graph into a relation-centric ontology graph, where each relation is represented by a type-level relation signature. Based on this ontology graph, OPI predicts the answer type implied by the question and maps it to a set of answer-type-compatible final-hop relations, enabling bidirectional retrieval that combines topic-side prefix expansion with answer-side final-hop matching. OPI then refines the retrieved evidence and candidate answers through a generator-refiner loop, where structured feedback iteratively updates both the path context and the answer context. \autoref{fig:overview_part2} summarizes the main inference pipeline of OPI.

\subsection{Ontology Graph Construction}
\label{sec:ontg-construct}


In large-scale knowledge graphs, the entity-level graph is often dense and highly heterogeneous. As the hop count increases, unconstrained topic-centered expansion can easily produce a large number of noisy evidence paths, causing rapid search-space growth. This motivates us to construct a compact type-level abstraction of the knowledge graph, so that repetitive entity-level facts can be summarized into more stable relation-type patterns.

Specifically, we abstract the factual knowledge graph $\mathcal{G}$ into a relation-centric ontology graph $\mathcal{O}$. In this graph, entity-level facts are summarized by type-level relation signatures of the form $(c_h, r, c_t)$, where $c_h$ and $c_t$ denote the head and tail entity types associated with relation $r$, respectively. Rather than preserving all concrete entity-level triples, the ontology graph captures how relations connect semantic types at the schema level. Since different knowledge bases expose type information in different forms, we adopt two construction strategies: schema-based extraction for knowledge bases with explicit schema predicates, and data-driven induction for knowledge bases without explicit schema predicates.

\subsubsection{For Freebase-style Knowledge Bases.}
When explicit schema predicates are available, relation signatures can be extracted directly from the schema. Specifically, we construct the ontology graph from the Freebase RDF dump~\cite{freebaserdf} under the canonical namespace \nolinkurl{http://rdf.freebase.com/ns/}. For each relation, we pair \texttt{type.property.\allowbreak schema} with \texttt{type.property.\allowbreak expected\_type} to obtain the head and tail types in its relation signature. The former specifies the domain type of the relation, while the latter specifies its expected range type. In this way, each schema-defined relation can be transformed into a type-level triple $(c_h, r, c_t)$.

To improve the reliability of the resulting ontology graph, we filter out non-semantic or administrative types, such as \texttt{common.topic}, and retain only relations with complete head-tail signatures. For a small number of missing or under-specified relations, we perform conservative completion using only the training split to avoid information leakage. Specifically, we aggregate the observed head and tail entity sets of each such relation from the training data, infer their dominant types, and accept the completion only when both sides admit a consistent type assignment. This strategy leverages the explicit schema structure of Freebase while improving coverage for rare or incomplete relations. The same construction principle naturally extends to other KGs with explicit schema predicates, such as DBpedia, where \texttt{rdfs:domain} and \texttt{rdfs:range} provide analogous type constraints.



\subsubsection{For Wiki-Movie-style Knowledge Bases.}
When explicit schema predicates are unavailable, relation signatures are induced directly from data. In Wiki-Movie-style KGs, the original graph does not provide schema predicates that explicitly define the domain and range types of each relation. Therefore, we first infer entity types from the training QA pairs and their associated type-path annotations. Each entity is assigned a single main type according to its most frequent observation, which reduces type sparsity and avoids unstable relation signatures caused by overly fine-grained or inconsistent type assignments.

After obtaining entity-level type assignments, we aggregate the observed head-tail type pairs for each relation. For a relation $r$, all training triples containing $r$ are mapped from entity-level triples $(e_h,r,e_t)$ to type-level observations $(c_h,r,c_t)$. We then retain the dominant head-tail type pair as the relation signature of $r$. This data-driven strategy allows ontology graph construction even when no explicit schema is available. More generally, for KGs such as Wikidata, relation signatures can also be induced by mapping entities to type sets via \texttt{P31} (\emph{instance of}) and optionally generalizing them with \texttt{P279} (\emph{subclass of}), followed by relation-wise aggregation of head and tail type statistics.

In both settings, the final ontology graph takes the same form: each relation is associated with a relation signature $(c_h, r, c_t)$. This unified relation-signature interface makes OPI applicable across heterogeneous knowledge graphs. More importantly, it provides the structural link from ontology construction to retrieval: in the next stage, predicted answer types are mapped to answer-type-compatible final-hop relations through exactly this interface.

\subsection{Ontology-Guided Bidirectional Retrieval}
Building on the relation-signature interface defined above, OPI introduces an ontology-guided bidirectional retrieval mechanism to alleviate path explosion. Here, \textit{bidirectional} does not refer to conventional two-sided entity-level BFS from the topic entity and concrete answer entities. Instead, it means that retrieval is jointly constrained by topic-side prefix expansion and answer-side type constraints. Specifically, OPI first predicts the answer type implied by the question, maps it to answer-type-compatible final-hop relations through the ontology graph, and then combines topic-side prefix expansion with answer-side final-hop matching to retrieve evidence paths and candidate answers.

\subsubsection{Answer-Type Prediction}
The relation-signature interface makes answer-type prediction a natural target for retrieval guidance. Compared with direct answer-entity prediction, answer-type prediction is substantially more compact and stable: questions with different surface forms and different gold answers often still share the same answer-side semantic category, such as \emph{person}, \emph{film}, or \emph{location}. Predicting the answer type before graph retrieval therefore provides a coarse but reliable target that can later be translated into structural constraints on the final hop.

To construct supervision, given a question $q$ with topic entity $e_q$ and gold answer $e_a$, we extract the shortest factual path
$
w_{z}(e_q,e_a)=(
e_q \xrightarrow{r_1} \cdots \xrightarrow{r_l} e_a)
$
with relation sequence $z=(r_1,\ldots,r_l)$ in $\mathcal{G}$. We use this path only as a minimal supervision signal. In particular, we take its last-hop relation $r_l$ and query the ontology graph $\mathcal{O}$ for relation signatures of the form $(c_h,r_l,c_a)$, where $c_h$ is a head type and $c_a$ is a tail type. The tail type in such a signature is then regarded as a valid answer entity type for the question. This yields the target set 
\begin{equation}
\label{eq:answer_type_set}
\mathcal{C}_a=\left\{c_a \;:\; \exists\, c_h,\ (c_h,r_l,c_a)\in\mathcal{S}\right\},
\end{equation}
If multiple ontology-consistent answer entity types are induced, we treat them as equally valid and define the supervision target as
\begin{equation}
\label{eq:answer_type_posterior}
Q(c_a \mid q, e_q, e_a,\mathcal{G}, \mathcal{O}) =
\begin{cases}
\displaystyle \frac{1}{|\mathcal{C}_a|}, & c_a \in \mathcal{C}_a,\\[4pt]
0, & \text{otherwise}.
\end{cases}
\end{equation}
Following \citet{luo2024rog}, we use this simple target distribution to supervise answer-type prediction.

We then fine-tune a large language model to generate the answer entity type conditioned on the question $q$. This defines a conditional prior
\begin{equation}
\label{eq:answer_type_prior}
\log P_{\phi}(c_a \mid q)
=
\sum_{i=1}^{|c_a|}
\log P_{\phi}(s_i \mid s_{<i}, q).
\end{equation}
where $\phi$ denotes the model parameters, $s_i$ is the $i$-th generated token, and $s_{<i}$ denotes the token prefix before position $i$. The resulting prediction is subsequently mapped to answer-type-compatible final-hop relations for retrieval.

\subsubsection{Answer-Type-Guided Bidirectional Retrieval}
Given a predicted answer type $c_a$, OPI uses the ontology graph to identify which relations can end at entities of this type. Specifically, we search the relation-signature set $\mathcal{S}$ for all signatures whose tail type is $c_a$. For each matched signature $(c_h,r,c_a)$, the relation $r$ is regarded as a candidate final-hop relation, because it can connect some head type $c_h$ to the predicted answer type $c_a$ at the ontology level. Formally, we define the answer-type-compatible final-hop relation set as
\begin{equation}
\label{eq:last_hop_relation_set}
\mathcal{R}^{\mathrm{last}}_{c_a}
=
\left\{
r \;:\; \exists\, c_h,\ (c_h,r,c_a)\in\mathcal{S}
\right\},
\end{equation}
Through this relation-signature interface, the predicted answer type is converted into a set of structurally plausible final-hop relations for subsequent retrieval.

Retrieval then proceeds bidirectionally. On the forward side, we start from the topic entity and traverse the factual graph along the original edge directions, expanding candidate path prefixes hop by hop under bounded depth and path budgets. On the answer side, rather than starting from concrete answer entities, OPI represents the answer constraint by a set of ontology-compatible final-hop relations $\mathcal{R}^{\mathrm{last}}_{c_a}$ induced by the predicted answer type. Importantly, this set is defined at the ontology level and may still contain multiple admissible relations for the target tail type. However, under a specific question, the feasible last hop is further constrained by the endpoint of the forward prefix. Concretely, for each prefix ending at a penultimate node $v$, only those relations in $\mathcal{R}^{\mathrm{last}}_{c_a}$ for which the type of $v$ matches the required head-side signature remain valid. Therefore, many ontology-legitimate final hops are eliminated immediately once the forward context is fixed, and the candidate set at the last step is typically reduced to only a few relations, and in many cases to a unique one.

The two sides thus meet at the penultimate node. For each forward prefix ending at $v$, we check whether $v$ can serve as a valid head entity for some $r \in \mathcal{R}^{\mathrm{last}}_{c_a}$. If so, we append the matched final hop and obtain a complete evidence path. Rather than expanding all outgoing relations at the final step, the search explicitly reserves the last hop for ontology-guided completion. This design not only enforces answer-type consistency, but also avoids introducing a large number of spurious branches that would otherwise arise from unconstrained topic-centered expansion.


This mechanism constrains the search space and reduces search complexity. Let $b$ denote the average branching factor of the factual knowledge graph. Without answer-side constraints, an unconstrained $x$-hop forward expansion from the topic entity explores
\begin{equation}
\label{eq:full_search_complexity}
|\mathcal{P}^{\mathrm{full}}_x(e_q)| = O(b^x).
\end{equation}

With answer-type-guided final-hop constraints, retrieval is reformulated as searching for valid $(x-1)$-hop prefixes followed by a constrained one-hop completion. Let $\mathcal{R}^{\mathrm{last}}_{c_a}$ denote the set of final-hop relations compatible with the predicted answer type $c_a$, and let $\beta(c_a)$ denote the effective branching factor after applying these final-hop constraints at prefix endpoints. Since the constrained completion is selected from the original outgoing branches, we have $\beta(c_a) \leq b$. The resulting search space is therefore bounded by
\begin{equation}
\label{eq:bidirectional_search_complexity}
|\mathcal{P}^{\mathrm{bi}}_x(e_q,c_a)| =
O\!\left(b^{x-1}\cdot \beta(c_a)\right).
\end{equation}
This formulation shows that OPI reduces the last-hop expansion from unconstrained entity-level branching to answer-type-constrained completion. The reduction depends on the selectivity of the answer-side constraints, rather than assuming that all ontology-compatible relation sets are uniformly small.

\subsubsection{Fallback Retrieval and Evidence Reranking}
If no answer type is predicted, or if the predicted answer type cannot be mapped to any answer-type-compatible final-hop relation through the relation-signature interface, we do not perform answer-side constrained matching. Instead, we revert to topic-centered forward retrieval from the topic entity and enumerate candidate paths up to a bounded number of hops. This fallback preserves a graph-grounded evidence source even when answer-type guidance is unavailable.

After candidate paths are obtained, we retain the top-$k$ evidence paths under a fixed path budget. In the current implementation, the candidate paths and the question are encoded by a pretrained language model (e.g., SentenceBERT~\cite{reimers2019sentence}), and their semantic relevance is measured in the embedding space. The top-$k$ paths are then retained and converted into readable path strings. Candidate answers are finally extracted from the tail entities of the retained paths, ensuring that the final answer space remains grounded in explicit graph evidence.

\subsection{Iterative Answer Refinement}
\label{sec:iterative-refine}
After ontology-guided bidirectional retrieval, OPI obtains a bounded set of candidate evidence paths and candidate answers extracted from their tail entities. Although this retrieval stage substantially reduces the search space, the remaining paths may still contain competing branches, partially relevant evidence, or semantically incomplete support for the final answer. OPI therefore does not directly return the top-ranked candidate. Instead, it performs iterative answer refinement through a generator-refiner loop.

\subsubsection{Generator-Refiner Loop}
At iteration $t$, the generator produces an answer hypothesis from the question $q$ and the current retrieval context. In the current implementation, this context includes the predicted answer-type constraints $c_a$, the reasoning-path context $P^{(t-1)}$, the answer context $A^{(t-1)}$, and the refinement feedback from the previous round $F^{(t-1)}$. Thus, the first iteration is already grounded in retrieved evidence, while later iterations are further constrained by revision signals.

Formally, the generation step at iteration $t$ is defined as
\begin{equation}
\label{eq:generator_refiner_loop_cr}
y^{(t)} = G_{\theta}(q, c_a, P^{(t-1)}, A^{(t-1)}, F^{(t-1)}),
\end{equation}
where $y^{(t)}$ is the current answer hypothesis. OPI further introduces a refiner that takes the current answer hypothesis together with a refinement-specific evidence subset and converts it into explicit revision actions:
\begin{equation}
\label{eq:refiner_interface_cr}
F^{(t)} = R_{\psi}(q, c_a, \bar{P}^{(t)}, y^{(t)}),
\end{equation}
where $\bar{P}^{(t)}$ is the refinement-specific path subset selected from the initial reranked path pool according to the current answer hypothesis. The refiner output includes a confidence score, an issue tag, retained answers, forbidden answers, prioritized paths, supplementary paths, dropped paths, and a short feedback summary. The generator proposes answer hypotheses, whereas the refiner converts the current answer state into structured revision actions.

\subsubsection{Refiner-Guided Context Update}
Starting from the second iteration, OPI no longer directly reuses the full retrieved path set in the next generator call. Instead, it reconstructs the next-round path context from the initial reranked path pool and the previous refinement result. Paths aligned with retained answers are preferred. If no retained answers are available, the system falls back to paths aligned with the previous generated answers. When the refiner indicates conflict or answer-set noise, OPI supplements the focused paths with a small number of neutral paths. Prioritized paths are promoted, supplementary paths are appended, and dropped paths are removed.

Formally, the next-round context is updated as
\begin{align}
\label{eq:path_context_update}
P^{(t)} &= \mathrm{Update}_{P}\bigl(P^{(0)}, y^{(t)}, F^{(t)}\bigr),\\
\label{eq:answer_context_update}
A^{(t)} &= \mathrm{Update}_{A}\bigl(y^{(t)}, F^{(t)}\bigr).
\end{align}
where $P^{(0)}$ denotes the initial reranked path pool. In the current implementation, $A^{(t)}$ corresponds to the retained answers exposed to the next generator round, while forbidden answers are injected separately through refinement feedback and explicit type-level answer constraints. Thus, the next-round path context is reconstructed from the initial path pool under refinement guidance, whereas the next-round answer context is formed by the supported answers retained by the refiner. If the stopping criterion is met, this context update is skipped.

\subsubsection{Iterative Revision and Stopping}
OPI adopts an adaptive stopping strategy based on two signals: refiner confidence and answer stability. The refiner outputs a discrete confidence level together with structured revision actions. The refinement loop terminates when the refiner assigns the highest confidence level, indicating that the current answer is sufficiently supported by the retrieved evidence. We use this highest confidence level as a conservative stopping threshold to avoid premature termination when the evidence remains ambiguous.

OPI also stops when the post-refinement answer remains unchanged across consecutive iterations. This stability criterion captures cases where further refinement no longer changes the final prediction. If either stopping condition is satisfied, OPI returns the answer from the current refinement round. Otherwise, the updated path context, answer context, answer-type constraints, and refinement feedback are passed to the next generator round. If no early stopping condition is triggered, OPI returns the answer from the final refinement round.

Overall, OPI closes the loop between retrieved evidence and final answer prediction. The generator proposes answer hypotheses under graph-grounded retrieval context, while the refiner converts the current answer state into explicit revision actions. The path and answer contexts are then updated accordingly for the next round. This adaptive process helps suppress noisy branches and improves answer reliability beyond single-pass retrieval-and-readout.

\begin{algorithm}[t]
\caption{OPI Inference Procedure}
\label{alg:OPI}
\small
\begin{algorithmic}[1]
\Statex \textbf{Input:} Question $q$, topic entity $e_q$, knowledge graph $\mathcal{G}$, ontology graph $\mathcal{O}$, path budget $k$, maximum refinement rounds $T$
\Statex \textbf{Output:} Final answer $\hat{y}$

\State \textbf{Step 1: Ontology-Guided Bidirectional Retrieval}
\State Predict answer type: $c_a \gets \Call{PredictType}{q}$
\State Map $c_a$ to valid final-hop relations: $\mathcal{R}^{\mathrm{last}}_{c_a} \gets \Call{MapToLastHop}{c_a,\mathcal{O}}$
\If{$\mathcal{R}^{\mathrm{last}}_{c_a} \neq \emptyset$}
    \State Retrieve candidate evidence paths by bidirectional search:
    \State \hspace{1em} $P^{(0)} \gets \Call{BidirectionalRetrieve}{e_q,\mathcal{R}^{\mathrm{last}}_{c_a},\mathcal{G}}$
\Else
    \State Retrieve candidate evidence paths by fallback search:
    \State \hspace{1em} $P^{(0)} \gets \Call{FallbackRetrieve}{e_q,\mathcal{G}}$
\EndIf
\State Rerank and retain the top-$k$ paths:
\State \hspace{1em} $P^{(0)} \gets \Call{RerankTopK}{q,P^{(0)},k}$
\State Extract initial answer candidates:
\State \hspace{1em} $A^{(0)} \gets \Call{ExtractAnswers}{P^{(0)}}$

\State \textbf{Step 2: Iterative Answer Refinement}
\State Initialize refinement feedback: $F^{(0)} \gets \emptyset$
\For{$t = 1$ to $T$}
    \State Generate the current answer hypothesis:
    \State \hspace{1em} $y^{(t)} \gets \Call{Generate}{q,c_a,P^{(t-1)},A^{(t-1)},F^{(t-1)}}$
    \State Select the refinement-specific path subset:
    \State \hspace{1em} $\bar{P}^{(t)} \gets \Call{SelectRefinePaths}{P^{(0)},y^{(t)}}$
    \State Refine the current answer and obtain revision actions:
    \State \hspace{1em} $F^{(t)} \gets \Call{Refine}{q,c_a,\bar{P}^{(t)},y^{(t)}}$
    \State Derive the refined answer for the current round:
    \State \hspace{1em} ${y}^{(t)} \gets \Call{PostRefine}{y^{(t)},F^{(t)}}$
    \If{$\Call{ShouldStop}{F^{(t)},{y}^{(t)}}$}
        \State \textbf{break}
    \EndIf
    \State Update the next-round path context and answer context:
    \State \hspace{1em} $P^{(t)} \gets \Call{UpdatePathContext}{P^{(0)},y^{(t)},F^{(t)}}$
    \State \hspace{1em} $A^{(t)} \gets \Call{UpdateAnswerContext}{y^{(t)},F^{(t)}}$
\EndFor

\State \Return $\hat{y} \leftarrow {y}^{(t)}$
\end{algorithmic}
\end{algorithm}

\section{Experiments}
We conduct extensive experiments to answer the following research questions:
\textbf{RQ1}: Does OPI outperform existing KGQA methods on WebQSP and CWQ?
\textbf{RQ2}: Is ontology-guided bidirectional retrieval effective as an independent retrieval module?
\textbf{RQ3}: How much does each component of OPI contribute to the overall performance?
\textbf{RQ4}: Why do ontology-based constraints improve retrieval effectiveness?
\textbf{RQ5}: Is OPI robust across different LLM backbones?
\textbf{RQ6}: Can OPI reduce retrieval cost and evidence noise?

\subsection{Experimental Settings}
\textbf{Datasets.}
We evaluate our method on three widely used KGQA benchmarks: WebQuestionsSP (WebQSP)~\citep{yih2016webqsp}, ComplexWebQuestions (CWQ)~\citep{talmor2018cwq}, and MetaQA~\citep{zhang2018variational}. These benchmarks are built on two knowledge graphs and cover both open-domain and domain-specific QA scenarios. Specifically, WebQSP and CWQ are based on Freebase~\citep{bollacker2008freebase}, while MetaQA is constructed on Wiki-Movie~\citep{miller2016key}. As shown in \autoref{tab:dataset_statistics}, they vary in dataset size, reasoning depth, and question complexity, providing a comprehensive evaluation setting for multi-hop KGQA.

For the Freebase setting, we evaluate OPI on WebQSP and CWQ, two standard benchmarks based on the preprocessed Freebase subgraphs widely used in prior work~\citep{he2021NSM,luo2024rog}. OPI constructs the relation-centric ontology graph from the full Freebase dump for more complete type-level constraints, while restricting retrieval and evaluation to the benchmark subgraphs for fair comparison. WebQSP contains 4,737 questions and mainly involves relatively simple reasoning, with answers typically located within two hops of the topic entity. CWQ is more challenging, containing 34,699 questions with compositional structures and additional constraints, often requiring up to four hops of reasoning. We follow the standard data splits for both datasets~\citep{sun2018open}.

For the Wiki-Movie setting, we adopt MetaQA, a movie-domain KGQA benchmark built on the Wiki-Movie knowledge graph, which contains 43,234 entities, 9 relations, and 133,582 triples. MetaQA includes more than 400K questions and is divided into three subsets by reasoning depth: MetaQA-1hop, MetaQA-2hop, and MetaQA-3hop. Following prior work~\cite{he2021NSM}, we evaluate on all three subsets and construct the one-shot setting by randomly sampling one training instance for each question template, resulting in 161, 210, and 150 training examples for the three subsets, respectively.




\begin{table}[hb]
\caption{Statistics of the experiment datasets.}
\label{tab:dataset_statistics}
\resizebox{\columnwidth}{!}{
\begin{tabular}{cccccc}
\toprule
Datasets & KG & \#Train & \#Dev & \#Test & Max \#Hops   \\
\midrule
WebQSP & Freebase & 2,826  & 246 & 1628  & 2  \\
CWQ    & Freebase & 27,639 & 3,519 & 3,531 & 4  \\
MetaQA-1hop & Wiki-Movie & 96,106 & 9,992 & 9,947 & 1   \\
MetaQA-2hop & Wiki-Movie & 118,980 & 14,872 & 14,872 & 2 \\
MetaQA-3hop & Wiki-Movie & 114,196 & 14,274 & 14,274 & 3 \\
\bottomrule
\end{tabular}
}
\end{table}

\textbf{Evaluation Metrics.}
Following prior work~\cite{sun2018open,luo2024rog,sun2024tog}, we use Hit@1 and F1 as the primary evaluation metrics. Hit@1 measures whether the top-ranked prediction matches any gold answer, reflecting top-1 answer accuracy. F1 evaluates the predicted answer set by jointly considering precision and recall, which is important for questions with multiple correct answers. In the ablation studies, we additionally report precision and recall for a more fine-grained analysis of prediction quality.

\textbf{Comparison Baselines.} We compare OPI with representative KGQA methods from four categories: embedding-based methods, retrieval-based methods, standalone LLMs, and KG-enhanced LLM methods. The first two categories cover conventional KG reasoning models that learn KG representations or retrieve question-relevant subgraphs, while standalone LLMs evaluate reasoning with parametric knowledge alone. We further include recent KG-enhanced LLM methods, such as ToG, RoG, ORT, and GCR, which are the most direct competitors to OPI. For MetaQA, we additionally compare with representative methods reported in prior work to evaluate retrieval-only multi-hop structural reasoning.

\textbf{Implementation Details.}
OPI uses LLaMA2-Chat-7B as the fine-tuned backbone LLM for answer-type prediction. The model is instruction-tuned for three epochs on four A100-40G GPUs with batch size 4, learning rate 2e-5, cosine learning rate scheduling, and a warmup ratio of 0.03. The joint training time is about 4.7 hours for WebQSP and CWQ, 44.0 hours for the MetaQA 1--3 hop data, and 0.42 hours for the MetaQA-oneshot setting. For ontology-guided bidirectional retrieval, we set the maximum reasoning depth according to dataset hop statistics, using two hops for WebQSP and four hops for CWQ. Retrieved evidence paths are ranked by Sentence-BERT similarity between the question and the path text, with \texttt{all-mpnet-base-v2} as the ranking model, and at most 256 evidence paths are retained for each question. For answer refinement, we use prompt-based LLMs with temperature 0.2, maximum output length 128, and one sampled response per question. Since different prompt-based models may exhibit different refinement behaviors, the number of refinement rounds is selected on the validation set: three rounds for DeepSeek-v3 and one round for GPT-4o. In all cases, OPI allows at most three refinement rounds and stops early when the refiner reaches high confidence or when two consecutive rounds produce stable answers.

\subsection{Overall Evaluation (RQ1)}
\autoref{tab:webqsp-cwq-results} reports the main results on WebQSP and CWQ. We compare OPI with four groups of representative methods, including embedding-based methods, retrieval-based methods, standalone LLMs, and recent KG-enhanced LLM methods. Overall, OPI achieves the best metric-wise performance on both datasets. Taking the best score among the two OPI variants for each metric, OPI achieves 92.3 Hit@1 and 76.8 F1 on WebQSP. Compared with the strongest previous results, it improves Hit@1 by 4.6 points and F1 by 5.0 points over ORT. On the more complex CWQ dataset, OPI achieves 76.5 Hit@1 and 62.7 F1, outperforming the metric-wise strongest prior results by 8.9 points in Hit@1 and 3.3 points in F1. These gains show that OPI is effective not only on WebQSP questions, which are generally less compositional, but also on complex questions in CWQ that require more careful multi-hop evidence retrieval.

\begin{table*}[t]
\caption{Comparison of methods on WebQSP and CWQ. The backbone model used by each method is shown in parentheses.}
\label{tab:webqsp-cwq-results}
\centering
\begin{tabular}{l l cc cc}
\toprule
\multirow{2}{*}{\textbf{Type}} & \multirow{2}{*}{\textbf{Methods}} & \multicolumn{2}{c}{\textbf{WebQSP}} & \multicolumn{2}{c}{\textbf{CWQ}} \\
 &  & \textbf{Hit@1} & \textbf{F1} & \textbf{Hit@1} & \textbf{F1} \\
\midrule
\multirow{4}{*}{Embedding-based}
& KV-Mem \cite{miller2016key}                              & 46.7 & 34.5 & 18.4 & 15.7 \\
& NSM \cite{he2021NSM}                              & 68.7 & 62.8 & 47.6 & 42.4 \\
& TransferNet \cite{shi2021transfernet}                              & 71.4 & \textemdash & 48.6 & \textemdash \\
& KGT5 \cite{saxena2022sequence}                              & 56.1 & \textemdash & 36.5 & \textemdash \\
\midrule
\multirow{4}{*}{Retrieval-based}
& GraftNet \cite{sun2018open}                       & 66.7 & 62.4 & 36.8 & 32.7 \\
& PullNet \cite{sun2019pullnet}                        & 68.1 & \textemdash & 45.9 & \textemdash \\
& SR+NSM \cite{zhang2022subgraph}                        & 68.9 & 64.1 & 50.2 & 47.1 \\
& UniKGQA \cite{jiang2023unikgqa}                        & 77.2 & 72.2 & 51.2 & 49.1 \\
\midrule
\multirow{6}{*}{LLMs}
& Llama-2-7B \cite{touvron2023llama}                  & 56.4 & 36.5 & 28.4 & 21.4 \\
& Llama-3.1-8B \cite{Meta2024llama3}                      & 55.5 & 34.8 & 28.1 & 22.4 \\
& ChatGPT \cite{openai2022chatgpt}                           & 59.3 & 43.5 & 34.7 & 30.2 \\
& GPT-4o \cite{openai2024gpt4o}                        & 61.8 & 43.6 & 38.2 & 32.9 \\
& DeepSeek-v3 \cite{liu2024deepseek}                        & 64.0 & 43.9 & 41.1 & 33.8 \\
\midrule
\multirow{8}{*}{KGs+LLMs}
& ToG (GPT-4) \cite{sun2024tog}                    & 82.6 & \textemdash & 67.6 & \textemdash \\
& RoG (Llama-2-7B) \cite{luo2024rog}                & 85.7 & 70.8 & 62.6 & 56.2 \\
& SymAgent (Llama-2-7B) \cite{liu2025symagent} & 55.5 & 41.3 & 35.1 & 31.2 \\
& GNN-RAG (Llama-2-7B) \cite{mavromatis2025gnnrag}              & 85.7 & 71.3 & 66.8 & 59.4 \\
& $R^2$ (Llama-2-7B) \cite{yuan2026reliable}              & 87.2 & 72.4 & 64.0 & 58.0 \\
& ORT (GPT-4o) \cite{liu2025ort}         & 87.7 & 71.8 & 65.4 & 58.7\\
& GCR (Llama-2-7B + GPT-4o) \cite{luo2025gcr}          & 87.5 & 71.5 & 66.0 & 58.5 \\
& \textbf{OPI(Llama-2-7B + GPT-4o)}           & 91.3 & \textbf{76.8} & 72.3 & \textbf{62.7} \\
& \textbf{OPI(Llama-2-7B + DeepSeek-v3)} & \textbf{92.3} & 74.9 & \textbf{76.5} & 59.6
\\
\bottomrule
\end{tabular}
\end{table*}

Beyond the overall performance, we further analyze the behavior of different method categories. Traditional embedding-based and retrieval-based methods can exploit KG structure through learned representations or retrieved subgraphs, but they still struggle on compositionally complex questions, as reflected by their consistently lower scores on CWQ. Standalone LLMs achieve reasonable results without explicit graph retrieval, but they generally lag behind KG-enhanced LLM methods, indicating that parametric knowledge alone is insufficient for reliable multi-hop KGQA. Recent KGs+LLMs methods further improve performance by combining LLM reasoning with graph evidence, yet many of them still retrieve evidence mainly from the topic entity side. As the hop number increases, such topic-centered expansion can introduce many type-mixed paths and semantically ambiguous candidates. In contrast, OPI incorporates type-level answer-side constraints into bidirectional retrieval and further refines candidate answers through a generator-refiner loop, improving both evidence quality and answer selection.

We also observe different strengths between the two OPI variants. OPI with DeepSeek-v3 achieves the highest Hit@1 on both WebQSP and CWQ, suggesting stronger top-answer selection. OPI with GPT-4o obtains the best F1 on both datasets, suggesting better coverage and calibration when multiple candidate answers are involved. Despite this difference, both variants consistently outperform prior KG-enhanced LLM methods, suggesting that the improvement is not tied to a specific backbone model, but is largely attributable to OPI's ontology-guided bidirectional retrieval and iterative refinement framework.

\subsection{Effectiveness of Ontology-Guided Bidirectional Retrieval (RQ2)}
To further examine whether ontology-guided bidirectional retrieval can provide effective graph evidence by itself, we evaluate OPI under a pure retrieval-only setting. In this setting, candidate answers are directly extracted from the endpoints of retrieved evidence paths, without any subsequent LLM-based answer generation or iterative refinement.

\begin{table}[hb]
\caption{Retrieval-only comparison with reproduced RoG and GCR on WebQSP and CWQ.}
\label{tab:retrieval-only-results}
\centering
\begin{tabular}{lcccc}
\toprule
\multirow{2}{*}{\textbf{Methods}} 
& \multicolumn{2}{c}{\textbf{WebQSP}} 
& \multicolumn{2}{c}{\textbf{CWQ}}  \\
\cmidrule(lr){2-3} \cmidrule(lr){4-5}
& \textbf{Hit@1} & \textbf{F1} 
& \textbf{Hit@1} & \textbf{F1}  \\
\midrule
RoG-BR \cite{luo2024rog} & 76.97 & 53.84 & 52.56 & 21.50  \\
GCR-BR \cite{luo2025gcr} & 92.19 & \textbf{58.03} & 69.12 & \textbf{39.44}  \\
\textbf{OPI-BR} & \textbf{95.39} & 39.09 & \textbf{88.95} & 29.78 \\
\bottomrule
\end{tabular}
\end{table}

\textbf{Results on WebQSP and CWQ.}
\autoref{tab:retrieval-only-results} compares OPI with reproduced RoG and GCR variants under the same retrieval-only setting. OPI-BR achieves the highest Hit@1 on both datasets, reaching 95.39 on WebQSP and 88.95 on CWQ. The improvement is particularly clear on CWQ, where compositional questions make topic-centered expansion more likely to introduce noisy and type-mixed paths. These results indicate that answer-side ontology constraints can effectively restrict retrieval to structurally more plausible endpoints under complex multi-hop reasoning, thereby providing useful graph-grounded evidence for subsequent answer generation and refinement.

However, the F1 results show a different trend. On WebQSP, OPI-BR obtains an F1 of 39.09, lower than RoG-BR and GCR-BR. On CWQ, OPI-BR improves over RoG-BR in F1, but still lags behind GCR-BR. This is because RoG and GCR already incorporate question-level semantics during path construction: RoG prompts the LLM to generate relation paths, while GCR further constrains this process with graph structure. In contrast, OPI-BR in this experiment mainly applies answer-side ontology constraints to retrieve answer-type-compatible endpoints, but does not yet perform full question-aware path verification or answer-set refinement. As a result, its raw retrieved endpoints may still contain type-compatible but question-irrelevant entities, which limits F1. Overall, this comparison shows that ontology-guided bidirectional retrieval is particularly effective for top-answer reachability and search-space reduction, but answer-set completeness still benefits from the subsequent refinement stages.

\begin{table*}[t]
\caption{Comparison of methods on MetaQA. Results of prior methods are reported as in their original papers. Our results are reported with two decimal places for better distinction near saturation.}
\label{tab:metaqa-results}
\begin{tabular}{l cc cc cc}
\toprule
\multirow{2}{*}{\textbf{Methods}} 
& \multicolumn{2}{c}{\textbf{MetaQA-1hop}} 
& \multicolumn{2}{c}{\textbf{MetaQA-2hop}} 
& \multicolumn{2}{c}{\textbf{MetaQA-3hop}} \\
\cmidrule(lr){2-3} \cmidrule(lr){4-5} \cmidrule(lr){6-7}
& \textbf{Hit@1} & \textbf{F1} 
& \textbf{Hit@1} & \textbf{F1} 
& \textbf{Hit@1} & \textbf{F1} \\
\midrule
KV-Mem \cite{miller2016key} & 96.2 & \textemdash & 82.7 & \textemdash  & 48.9 & \textemdash \\
NSM \cite{he2021NSM} & 97.1 & \textemdash & 99.9 & \textemdash  & 98.9 & \textemdash \\
GraftNet \cite{sun2018open} & 97.0 & \textemdash & 94.8 & \textemdash  & 77.7 & \textemdash \\
UniKGQA \cite{jiang2023unikgqa} & 97.5 & \textemdash & 99.0 & \textemdash  & 99.1 & \textemdash \\
StructGPT \cite{jiang2023structgpt} & 97.1 & \textemdash & 97.3 & \textemdash  & 87.0 & \textemdash \\
ReasoningLM \cite{jiang2023reasoninglm} & 96.5 & \textemdash & 98.3 & \textemdash  & 92.7 & \textemdash \\
KG-GPT \cite{kim2023kg} & 96.3 & \textemdash & 94.4 & \textemdash  & 94.0 & \textemdash \\
Retrieval \& Reasoning \cite{ji2024retrieval} & \textemdash & \textemdash & \textemdash & \textemdash  & 76.0 & 33.8 \\
RoG \cite{luo2024rog} & \textemdash & \textemdash & \textemdash & \textemdash  & 84.8 & 41.3 \\
BYOKG \cite{agarwal2024bring} & 95.3 & \textemdash & 81.9 & \textemdash  & 75.7 & \textemdash \\
ARG \cite{zhang2025learning} & \textemdash & \textemdash & \textemdash & \textemdash  & 87.7 & \textemdash \\
SymAgent \cite{liu2025symagent} & \textemdash & \textemdash & \textemdash & \textemdash  & 57.0 & 25.8 \\
KG-Agent \cite{jiang2025kg} & 97.1 & \textemdash & 98.0 & \textemdash  & 92.1 & \textemdash \\
RDPG \cite{ding2025enhancing} & 99.7 & \textemdash & 98.6 & \textemdash  & 87.7 & \textemdash \\
$R^2$ \cite{yuan2026reliable} & \textemdash & \textemdash & \textemdash & \textemdash  & 85.2 & 42.5 \\
\textbf{OPI-BR} & \textbf{100.00} & \textbf{96.94} & \textbf{99.99} & \textbf{83.51} & \textbf{99.96} & \textbf{59.19} \\
\textbf{OPI-BR-oneshot} & 99.68 & 95.68 & 99.94 & 83.50 & 99.86 & 58.60 \\
\bottomrule
\end{tabular}
\end{table*}

\textbf{Results on MetaQA.}
We further evaluate the bidirectional retrieval module of OPI on MetaQA, where questions are more template-like and relation paths are relatively regular. Compared with WebQSP and CWQ, MetaQA relies more on structural path matching and less on complex semantic disambiguation, making it suitable for evaluating pure retrieval-only reasoning. Therefore, we report bidirectional retrieval results without additional LLM-based answer generation, because the retrieved endpoints already provide a direct test of whether the structural relation path can reach the correct answer. Here, BR denotes bidirectional retrieval and does not include any subsequent LLM-based answer generation.

\autoref{tab:metaqa-results} reports the results on the full MetaQA benchmark. OPI-BR achieves near-saturated Hit@1 across all three subsets, with 100.00 on 1-hop, 99.99 on 2-hop, and 99.96 on 3-hop questions. OPI-BR-oneshot obtains very close performance, especially on 2-hop and 3-hop questions. These results show that when evidence paths are regular and answer-side constraints are reliable, ontology-guided bidirectional retrieval alone can already provide sufficient evidence for accurate answer prediction.

\subsection{Ablation Study (RQ3)}
We conduct ablation studies to assess three key designs in OPI: type-level search space, early answer-side constraint, and iterative answer refinement. All variants use the same Llama-2-7B + DeepSeek-v3 setting. \autoref{tab:ablation} reports the results on WebQSP and CWQ in terms of Hit@1, F1, precision, and recall.

\begin{table*}[t]
\caption{Ablation study results on WebQSP and CWQ. All variants use the same Llama-2-7B + DeepSeek-v3 setting.}
  \label{tab:ablation}
  \begin{tabular}{lcccc|cccc}
    \toprule
    \multirow{2}{*}{\textbf{Method}} &
      \multicolumn{4}{c}{\textbf{WebQSP}} &
      \multicolumn{4}{c}{\textbf{CWQ}} \\
    \cmidrule(lr){2-5}\cmidrule(lr){6-9}
    & \textbf{Hit@1} & \textbf{F1} & \textbf{Precision} & \textbf{Recall} 
    & \textbf{Hit@1} & \textbf{F1} & \textbf{Precision} & \textbf{Recall} \\
    \midrule
    \textbf{OPI}  & 92.32 & \textbf{74.91} & \textbf{80.16} & 78.95 & 76.52 & 59.59 & 58.96 & 71.23\\
    w/o type-level search space  & 82.13 & 60.96 & 65.35 & 66.47 & 54.15 & 41.05 & 41.24 & 48.13\\
    w/o in-retrieval answer-side constraint  & 86.43 & 66.75 & 73.46 & 69.96 & 59.08 & 45.61 & 46.88 & 51.90\\
    w/o iterative answer refinement  & \textbf{93.43} & 70.95 & 70.04 & \textbf{84.96} & \textbf{81.28} & 56.42 & 52.32 & \textbf{78.06}\\
    \bottomrule
  \end{tabular}
\end{table*}

\textbf{Effect of type-level search space.}
This variant removes the type-level path space and directly traverses the original KG from the topic entity. When the number of outgoing edges is large, it keeps the top-$k$ paths according to the SentenceBERT similarity between each path and the question. This change leads to the largest performance drop: Hit@1/F1 decrease by 10.19/13.95 points on WebQSP and by 22.37/18.54 points on CWQ. The degradation suggests that direct topic-centered expansion is easily affected by large branching factors. Although similarity-based pruning reduces the search cost, it may discard correct paths before they reach the answer side. By contrast, OPI uses type-level relation signatures to form a more compact search space, which helps preserve answer-relevant paths under multi-hop expansion.

\textbf{Effect of in-retrieval answer-side constraint.}
This variant removes answer-side constraints from the retrieval process. Instead, it first performs topic-side forward expansion and then applies answer-type-compatible final-hop relations as a post-retrieval filter. F1 decreases from 74.91 to 66.75 on WebQSP and from 59.59 to 45.61 on CWQ, showing that answer-side constraints are more effective when they participate in retrieval rather than only filtering completed paths. Since relevant paths may already be pruned during topic-centered forward expansion, post-retrieval filtering cannot recover them. Nevertheless, this variant still outperforms \textit{w/o type-level search space}, because the delayed final-hop filter can still remove some answer-type-incompatible endpoints.

\textbf{Effect of iterative answer refinement.}
This variant directly uses the single-pass output of the same prompt-based LLM as the final answer, without applying generator-refiner iterations. It achieves higher Hit@1 and recall on both datasets, but lower precision and F1. For example, compared with full OPI on CWQ, recall increases from 71.23 to 78.06, while precision decreases from 58.96 to 52.32 and F1 decreases from 59.59 to 56.42. This indicates that single-pass generation tends to retain a broader answer set, but also introduces more false positives. Iterative refinement therefore does not simply maximize coverage; instead, it trades part of the aggressive candidate retention for a cleaner answer set, suggesting that refinement acts more as a precision-oriented filtering mechanism than as a recall-maximizing step.

\subsection{Analysis of Ontology Graph (RQ4)}

\begin{table*}[t]
  \caption{Statistics of ontology graph construction for Freebase and Wiki-Movie.}
  \label{tab:ontology-stats}
  \begin{tabular}{lll}
    \toprule
    \textbf{Item} & \textbf{Freebase} & \textbf{Wiki-Movie} \\
    \midrule
    Schema setting 
      & Explicit schema predicates 
      & No explicit schema predicates \\
      
    Input entries 
      & 71{,}210 schema entries 
      & 134{,}741 triples \\
      
    Total pipeline time (single machine)
      & $\approx$ 1.93 h 
      & $\approx$ 1.17 s \\
      
    Incomplete signatures in WebQSP
      & 19/5{,}726 (0.33\%) 
      & - \\
      
    Incomplete signatures in CWQ
      & 21/6{,}576 (0.32\%) 
      & - \\
      
    Shared incomplete signatures in WebQSP and CWQ
      & 19/6{,}837 (0.28\%)
      & - \\
      
    Incomplete signatures in MetaQA
      & - 
      & 0/9 (0.00\%) \\
      
    Relations in ontology graph
      & 32{,}195 
      & 9 \\
      
    Entity types in ontology graph
      & 12{,}369 
      & 11 \\
    \bottomrule
  \end{tabular}
\end{table*}

\autoref{tab:ontology-stats} reports the construction statistics of the ontology graphs for Freebase and Wiki-Movie. The results show that the proposed construction procedure is feasible under different schema settings. For Freebase, which provides explicit schema predicates, the pipeline processes 71{,}210 schema entries from the RDF dump and completes in approximately 6{,}958 seconds ($\approx$1.93 hours) on a single machine. For Wiki-Movie, where explicit schema predicates are unavailable, the ontology graph is induced directly from 134{,}741 triples and constructed in approximately 1.17 seconds. This contrast mainly reflects the much larger scale and richer schema structure of Freebase, while also showing that the same ontology abstraction can be instantiated efficiently for compact domain-specific KGs.

The constructed ontology graphs cover most relations used in downstream KGQA benchmarks. For WebQSP, only 19 out of 5{,}726 relations have missing or incomplete signatures, accounting for 0.33\%. For CWQ, this number is 21 out of 6{,}576, accounting for 0.32\%. Moreover, 19 missing or incomplete signatures are shared by WebQSP and CWQ, corresponding to only 0.28\% of the 6{,}837 unique relations across both datasets. These results show that the Freebase-derived ontology graph covers nearly all benchmark relations. For MetaQA, all 9 relations have valid head--tail type signatures under the induced Wiki-Movie ontology.

The resulting ontology graphs are also compact relative to their underlying knowledge bases. The Freebase ontology graph contains 32{,}195 relation signatures and 12{,}369 entity types, providing broad schema coverage without direct expansion over dense entity-level facts. The Wiki-Movie ontology graph is much smaller, with only 9 relation signatures and 11 entity types, yet still preserves the type-level constraints required by MetaQA. These statistics suggest that the ontology graph serves as a lightweight abstraction: it retains the type semantics needed for retrieval while substantially simplifying the structure used during multi-hop reasoning.

Overall, these results show that ontology graph construction is scalable and effective across heterogeneous KG settings. Explicit schema predicates provide reliable relation signatures for schema-rich KGs, while schema-light KGs can be handled through data-driven induction. Thus, the ontology graph offers a lightweight and practical type-level interface for retrieval in multi-hop KGQA.


\subsection{Effect of Different LLM Backbones (RQ5)}

\begin{table}[hb]
\caption{Performance of different LLM backbone variants on WebQSP and CWQ.}
\label{tab:model_variants}
\resizebox{\columnwidth}{!}{
\centering
\begin{tabular}{llcccc}
\toprule
\multirow{2}{*}[-0.6ex]{\shortstack{\textbf{Fine-tuned}\\\textbf{models}}} &
\multirow{2}{*}[-0.6ex]{\shortstack{\textbf{Prompt-based}\\\textbf{models}}} & 
\multicolumn{2}{c}{\textbf{WebQSP}} & 
\multicolumn{2}{c}{\textbf{CWQ}} \\
\cmidrule(r){3-4} \cmidrule(l){5-6}
 &  & \textbf{Hit@1} & \textbf{F1} & \textbf{Hit@1} & \textbf{F1}\\
\midrule
\multirow{2}{*}{\shortstack{Qwen2-1.5B}}
  & DeepSeek-v3  & 91.28 & 73.47 & 75.70 & 58.71 \\
  & GPT-4o       & 90.66 & 76.28 & 71.40 & 61.85 \\
\midrule
\multirow{2}{*}{\shortstack{Qwen2-7B}}
  & DeepSeek-v3  & 91.03 & 73.51 & 75.64 & 59.03 \\
  & GPT-4o       & 90.36 & 75.64 & 70.57 & 61.19 \\
\midrule
\multirow{2}{*}{\shortstack{Llama-2-7B}}
  & DeepSeek-v3  & 92.32 & 74.91 & 76.52 & 59.59 \\
  & GPT-4o       & 91.34 & 76.83 & 72.33 & 62.73 \\
\bottomrule
\end{tabular}
}
\end{table}

To examine whether OPI depends on a specific LLM backbone, we evaluate different combinations of fine-tuned models and prompt-based models. The fine-tuned models are task-adapted LLMs used for answer-type prediction, including Qwen2-1.5B, Qwen2-7B, and Llama-2-7B. The prompt-based models are instruction-following LLMs used in the answer refinement stage, including DeepSeek-v3 and GPT-4o. \autoref{tab:model_variants} reports the results on WebQSP and CWQ.

Overall, OPI performs consistently well across different model combinations. On WebQSP, all variants achieve over 90 Hit@1 and over 73 F1. On CWQ, they also maintain strong performance, with Hit@1 ranging from 70.57 to 76.52 and F1 ranging from 58.71 to 62.73. These results indicate that OPI is not tied to a single LLM choice. Instead, its effectiveness largely stems from the ontology-guided bidirectional retrieval and iterative answer refinement framework, which can be adapted to different fine-tuned and prompt-based models.

We further observe complementary behaviors between prompt-based models. Variants using DeepSeek-v3 generally achieve higher Hit@1, especially on CWQ, suggesting stronger top-answer selection. In contrast, variants using GPT-4o obtain higher F1 on both datasets, indicating better answer-set calibration when multiple candidate answers are involved. Among the fine-tuned models, Llama-2-7B achieves the best overall results under both prompt-based models. Nevertheless, Qwen2-1.5B and Qwen2-7B remain competitive, showing that OPI can also operate effectively with smaller models for answer-type prediction.

\subsection{Efficiency and Robustness Analysis (RQ6)}

\begin{figure*}
  \centering
  \includegraphics[width=\linewidth]{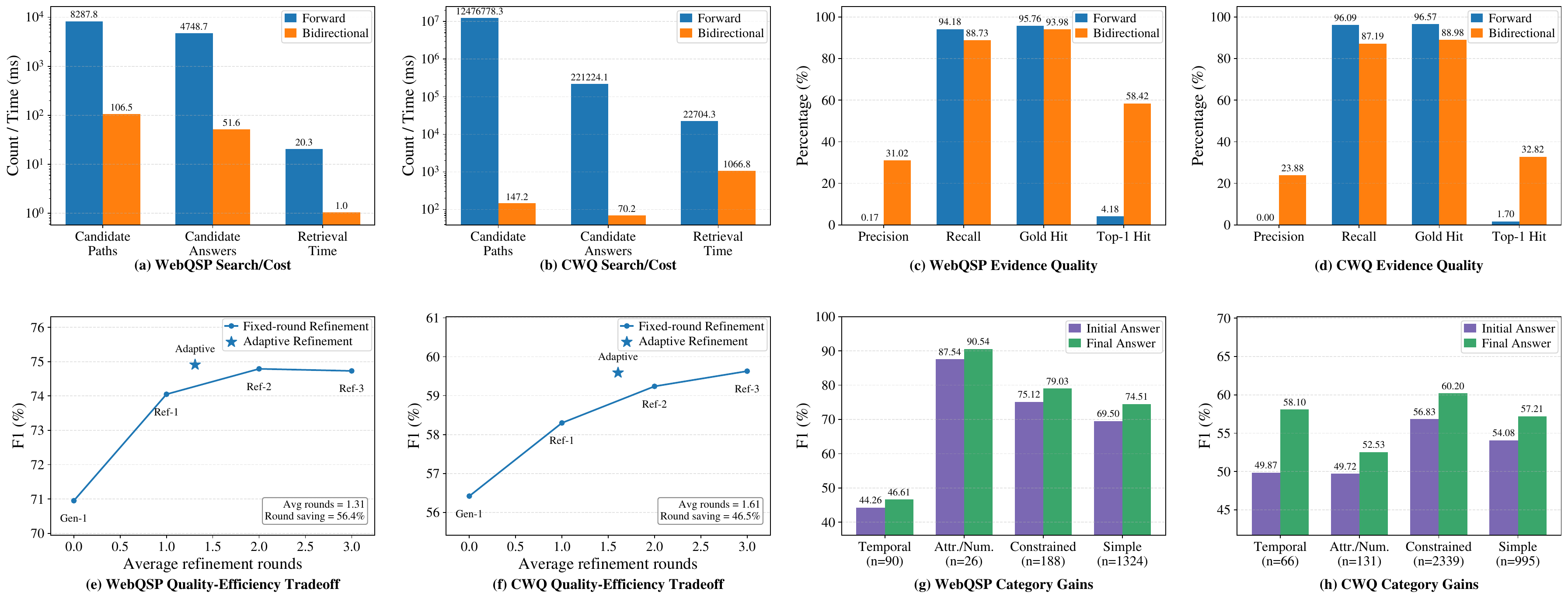}
    \caption{Efficiency and robustness analysis of OPI. 
    (a)-(b) Search-space and retrieval-cost comparison between forward-only and bidirectional retrieval. 
    (c)-(d) Evidence quality comparison before answer generation. 
    (e)-(f) Quality-efficiency tradeoff of adaptive refinement. 
    (g)-(h) F1 gains across question categories.
    }
  \label{fig:efficiency}
\end{figure*}


\textbf{Search-space and retrieval-cost reduction.}
We compare OPI's ontology-guided bidirectional retrieval with a forward-only baseline by measuring the average number of candidate paths, candidate answers, and retrieval time. As shown in Figures~\ref{fig:efficiency}(a) and~\ref{fig:efficiency}(b), OPI reduces candidate paths by 98.7\% on WebQSP and over 99\% on CWQ, candidate answers by 98.9\% and 99.97\%, and retrieval time by 95.1\% and 95.3\%, respectively. These reductions show that answer-side type constraints effectively prevent uncontrolled topic-centered expansion. By reserving the final hop for ontology-guided matching, OPI avoids exploring many type-incompatible evidence paths and therefore lowers retrieval cost.

\textbf{Evidence cleanliness and answer coverage.}
We evaluate the retrieved evidence before answer generation using precision, recall, gold-hit rate, and top-1 hit rate. Figures~\ref{fig:efficiency}(c) and~\ref{fig:efficiency}(d) show that OPI improves precision by 30.85\% on WebQSP and 23.88\% on CWQ, and improves top-1 hit rate by 54.24\% and 31.12\%, respectively. Although recall and gold-hit rate slightly decrease, they remain at 88.73\%/93.98\% on WebQSP and 87.19\%/88.98\% on CWQ. This indicates that OPI filters noisy evidence while preserving most answer-relevant candidates.

\textbf{Quality-efficiency tradeoff of adaptive refinement.}
We compare fixed-round refinement with adaptive refinement, where the refinement process stops based on answer stability and refiner confidence. As shown in Figures~\ref{fig:efficiency}(e) and~\ref{fig:efficiency}(f), adaptive refinement achieves 74.91 F1 with only 1.31 rounds on WebQSP, slightly outperforming fixed three-round refinement. On CWQ, it achieves 59.59 F1 with 1.61 rounds, closely matching the fixed three-round result. This reduces the average number of refinement rounds by 56.4\% and 46.5\%, respectively, showing that adaptive refinement avoids unnecessary iterations while maintaining answer quality.


\textbf{Benefits on semantically challenging questions.}
We compare the initial answer and the final refined answer across temporal, attribute/numerical, constrained, and simple questions. Figures~\ref{fig:efficiency}(g) and~\ref{fig:efficiency}(h) show consistent F1 gains across all categories. On WebQSP, the largest gain appears on simple questions, with a 5.01-point improvement, followed by constrained questions with a 3.91-point gain. On CWQ, temporal questions benefit the most, with an 8.23-point improvement, while constrained and simple questions improve by 3.37 and 3.13 points, respectively. These results suggest that iterative answer refinement helps OPI mitigate semantic ambiguity, complementing bidirectional retrieval, which primarily addresses structural path explosion.

\section{Related Work}
\label{sec:related-work}

\subsection{Multi-hop Knowledge Graph Question Answering}
Existing multi-hop KGQA methods can be broadly grouped into embedding-based methods, retrieval-based methods, standalone LLM methods, and KG-enhanced LLM methods. Early embedding-based methods, such as KV-Mem \citep{miller2016key}, NSM \citep{he2021NSM}, TransferNet \citep{shi2021transfernet}, and KGT5 \citep{saxena2022sequence}, learn graph-aware representations by encoding questions and entities in continuous spaces or propagating question-aware signals over graph structures. Although effective, their reasoning process is often implicit and may not preserve explicit multi-hop path semantics. Retrieval-based methods instead construct question-relevant subgraphs or evidence paths before answer prediction, as in GraftNet \citep{sun2018open}, SR+NSM \citep{zhang2022subgraph}, and UniKGQA \citep{jiang2023unikgqa}. While these methods improve interpretability by exposing supporting evidence, they typically rely on topic-centered expansion and may retrieve many structurally reachable but semantically irrelevant paths as reasoning depth increases.


With the emergence of large language models, standalone LLMs have been applied to KGQA through prompting, in-context learning, or chain-of-thought reasoning. Models such as Llama-2, Llama-3.1, ChatGPT, GPT-4o, and DeepSeek-v3 show strong language understanding and reasoning abilities \citep{touvron2023llama,Meta2024llama3,openai2022chatgpt,openai2024gpt4o,liu2024deepseek}, but they may hallucinate unsupported facts and are limited by the absence of explicit graph-grounded evidence. Recent KG-enhanced LLM methods address this limitation by incorporating retrieved triples, evidence paths, graph structures, or symbolic tools into LLM-based reasoning. For example, ToG explores reasoning chains over KGs with LLM guidance \citep{sun2024tog}, RoG generates relation paths for graph-grounded reasoning \citep{luo2024rog}, SymAgent combines symbolic reasoning with agentic LLM behaviors \citep{liu2025symagent}, GNN-RAG integrates graph neural retrieval with LLM generation \citep{mavromatis2025gnnrag}, and GCR further improves KG-enhanced reasoning through graph-context reasoning \citep{luo2025gcr}. Recent work such as $R^2$ also emphasizes reliable reasoning with LLMs over KGs \citep{yuan2026reliable}. These methods substantially improve graph-grounded answer generation, but their retrieval stages are still largely driven by topic-side exploration or post-retrieval evidence selection. In contrast, OPI introduces answer-side type constraints into the retrieval process itself, using a relation-centric ontology graph to guide final-hop matching and reduce noisy path explosion before LLM-based answer refinement.

\subsection{Ontology and Type-level Guidance}
Ontology and schema information provide high-level semantic structures for organizing entities, relations, types, and constraints in knowledge graphs. Compared with entity-level triples, type-level abstractions capture more stable semantic regularities, such as the expected head and tail types of relations. Recent studies have explored LLMs for ontology construction, enhancement, and learning \citep{funk2023towards,toro2024dynamic,babaei2023llms4ol}, as well as ontology construction from large RDF resources or capability-question-driven workflows \citep{chen2018research,kommineni2024human}. These studies show that ontology construction and enrichment can provide useful foundations for knowledge-intensive tasks.

In KGQA, ontology information has been used to improve the semantic alignment between natural-language questions and graph structures. For example, ontology-guided prompting and reverse-thinking strategies have been introduced to strengthen multi-hop reasoning and generalization \citep{jiang2025ontology,liu2025ort}, while OntoTune aligns LLMs with domain ontologies through ontology-driven self-training \citep{liu2025ontotune}. However, most existing methods use ontology information only as auxiliary signals, providing limited graph-grounded constraints in large heterogeneous retrieval spaces. Industrial systems such as Palantir further show that ontology can serve as a structured interface connecting data, reasoning, and downstream actions \citep{palantir2024aipontology}. OPI follows this motivation but specializes it for multi-hop KGQA: it constructs a relation-centric ontology graph from the KG itself and uses type-level signatures to map the predicted answer type to compatible final-hop relations, thereby integrating type-level and entity-level path search.

\section{Conclusion}

In this paper, we proposed OPI, an ontology-guided evidence path inference framework for multi-hop KGQA. OPI introduces a relation-centric ontology graph to make answer-side type constraints explicit, and combines topic-side prefix expansion with answer-side final-hop matching to reduce noisy path explosion. It further employs a generator-refiner loop to jointly reassess retrieved evidence and answer hypotheses, filtering type-compatible but question-irrelevant candidates. Experiments on WebQSP, CWQ, and MetaQA show that OPI consistently outperforms representative multi-hop KGQA methods. These results demonstrate that OPI provides a compact and reusable relation-centric ontology graph for retrieval, while effectively alleviating path explosion and semantic misalignment in multi-hop KGQA.

In future work, we plan to extend OPI to knowledge graphs with weaker or less reliable type information. A key direction is to construct and update relation signatures when entity types are missing, noisy, or only partially available, so that ontology-guided retrieval remains effective beyond KGs with well-defined schemas. Such an extension would broaden the applicability of OPI to open-domain and dynamically evolving knowledge graphs.








\bibliographystyle{ACM-Reference-Format}
\bibliography{sample}

\end{document}